\documentclass{vgtc}                          % final (conference style)
\ifpdf%                                % if we use pdflatex
  \pdfoutput=1\relax                   % create PDFs from pdfLaTeX
  \usepackage{graphicx}                % allow us to embed graphics files
  \DeclareGraphicsExtensions{.pdf,.png,.jpg,.jpeg} % for pdflatex we expect .pdf, .png, or .jpg files
\else%                                 % else we use pure latex
  \usepackage{graphicx}                % allow us to embed graphics files
  \DeclareGraphicsExtensions{.eps}     % for pure latex we expect eps files
\fi%

\graphicspath{{figures/}{pictures/}{images/}{./}} % where to search for the images

\usepackage{microtype}                 % use micro-typography (slightly more compact, better to read)
\PassOptionsToPackage{warn}{textcomp}  % to address font issues with \textrightarrow
\usepackage{textcomp}                  % use better special symbols
\usepackage{mathptmx}                  % use matching math font
\usepackage{times}                     % we use Times as the main font
\usepackage{cite}                      % needed to automatically sort the references
\usepackage{tabu}                      % only used for the table example
\usepackage{booktabs}                  % only used for the table example
\onlineid{2175}

\vgtccategory{Research}

\vgtcinsertpkg

\title{Multifaceted 4D Feature Segmentation and Extraction \\ in Point and Field-based Datasets}

\author{Franz Sauer\thanks{e-mail: fasauer@ucdavis.edu}\\ %
        \scriptsize University of California, Davis %
\and Kwan-Liu Ma\thanks{e-mail: ma@cs.ucdavis.edu}\\ %
     \scriptsize University of California, Davis}

\abstract{The use of large-scale multifaceted data is common in a wide variety of scientific applications. In many cases, this multifaceted data takes the form of a field-based (Eulerian) and point/trajectory-based (Lagrangian) representation as each has a unique set of advantages in characterizing a system of study. Furthermore, studying the increasing scale and complexity of these multifaceted datasets is limited by perceptual ability and available computational resources, necessitating sophisticated data reduction and feature extraction techniques. In this work, we present a new 4D feature segmentation/extraction scheme that can operate on both the field and point/trajectory data types simultaneously. The resulting features are time-varying data subsets that have both a field and point-based component, and were extracted based on underlying patterns from both data types. This enables researchers to better explore both the spatial and temporal interplay between the two data representations and study underlying phenomena from new perspectives. We parallelize our approach using GPU acceleration and apply it to real world multifaceted datasets to illustrate the types of features that can be extracted and explored.} % end of abstract

\keywords{Point data, field data, trajectories, segmentation, feature extraction, time-varying data}

\begin{document}


\firstsection{Introduction}

\maketitle

%% \section{Introduction} %for journal use above \firstsection{..} instead

%segmentation and feature extraction
The increasing size and complexity of scientific data is necessitating the use of more sophisticated data reduction and feature extraction techniques. This is due to both limits in the perceptual ability of researchers and limits in the storage and computing power of available hardware. By segmenting and/or extracting meaningful features from a dataset, scientists can focus their exploration in more manageable data subsets. However, the vast majority of segmentation and feature extraction techniques for scientific data focus on only one data representation at a time. This makes it difficult for researchers to draw connections across different data types and limits ways in which one can explore a system from multiple perspectives.

%multifaceted data examples
Multifaceted data is common in science, and even a single simulation can produce data in multiple formats. For instance, many combustion simulations~\cite{Yoo:2011} produce both particle and field data as each representation has a distinct set of advantages in characterizing the properties of such a system. Furthermore, data need not always be obtained from the same source, since different acquisition methods can also be used to study a particular system. For example, one can couple geospatial movement data of taxis (a point/trajectory-based representation) with measurements of traffic at discrete locations (a field-based representation)~\cite{Yu:2010,Zhang:2011}.

%point and field based representations
In many of these examples, the data often takes the form of field-based or point/trajectory-based representations. This is because each has a unique set of advantages in portraying system behavior. The field-based, or Eulerian, representation measures information at fixed spatial locations within a system domain. This allows one to easily capture properties, such as electromagnetic field strengths, temperature, or traffic patterns, in predefined unchanging locations. On the other hand, the point/trajectory-based, or Lagrangian, representation consists of quantities which are free to move throughout the system domain. This point data can be comprised of imaginary entities, such as massless tracers, or physical entities, such as electrons or motor vehicles.  Analyzing the geometric properties of trajectories allows one to capture motion and flow in intuitive ways.

%why we need both representations
As a result, new visualization and analysis techniques must incorporate both data representations in order for users to obtain a complete understanding of all aspects of the physical system. It is simply not feasible to project all information into one of the representations without losing the advantages of the other and can make analysis even more difficult. For example, trying to represent a particle as field data loses information about the motion of such an object. In addition, trying to sample field values onto the point data has no guarantee that there will be enough points in key locations within the domain. Instead, coupling both of these data types and allowing them to support one another can more effectively allow researchers to gain new insights about their system of study.

%what this paper is about
This work focuses on developing a feature segmentation and extraction scheme for datasets with both a point and field-based representation. By incorporating information from both of these data types, this technique can generate a meaningful set of 4D multifaceted features based on complex trends found within the data. Since these features contain both field and point-based components, and span multiple timesteps, users are able to explore spatio-temporal properties in detailed data subsets from the perspective of both data types simultaneously. Such a method allows for more control over the types of data subsets produced and can generate information rich features previously unavailable to researchers.

%challenges
This presents a number of challenges due to the fundamental differences of each representation. While the field-based representation has a strong spatial structure, features that span across timesteps are difficult to identify since they often involve discrete jumps between grid locations as they propagate throughout the domain. On the other hand, the point-based representation has a strong temporal coherence since trajectories naturally connect points between timesteps, but it lacks the spatial structure available in the field data. Another challenge lies in managing large data scales since an algorithm that can segment such a dataset into 4D features will need fast access to all timesteps of each data type. This necessitates the use of an algorithm that can work in parallel and/or in an out of core fashion.

%contributions
This paper describes our multifaceted feature segmentation and extraction technique for use in point-based and field-based datasets. Overall, we contribute an algorithm that is able to automatically segment such data into a unique set of multifaceted features that:

\vspace{-\topsep}
\begin{itemize} \itemsep1pt \parskip1pt \parsep1pt
  \item contain both point and field-based components
  \item were extracted based on a combination of patterns in both the point and field data types
  \item are 4D in nature (time-varying 3D)
\end{itemize}
\vspace{-\topsep}

\noindent We demonstrate our method through case studies using real world datasets and illustrate its ability to extract useful multifaceted features from the data.

% RELATED WORK %
%%%%%%%%%%%%%%%%%%%%%%%%%%%%%%%%%%%%%%%%%%%%%%%%%%
%%%%%%%%%%%%%%%%%%%%%%%%%%%%%%%%%%%%%%%%%%%%%%%%%%

\section{Background and Related Work} \label{sec:related_work}

%surveys
Although very little work has been done on extracting features that incorporate information from both point and field data simultaneously, there has been an extensive set of previous work focused on visualization and feature extraction separately in point-based or field-based datasets. A good overview can be found in the various survey papers related to this topic, such as the one by Kehrer and Hauser~\cite{Kehrer:2013}, which describes techniques for visualization of multifaceted scientific data, and Fuchs and Hauser~\cite{Fuchs:2009}, which focuses on visualization of multi-variate scientific data. A broader survey by Andrienko et al.~\cite{Andrienko:2003} describes general spatio-temporal visualization techniques outside the realm of scientific visualization. Another survey by Brambilla et al.~\cite{Brambilla:2012} describes illustrative flow rendering techniques, which involves methods of visualizing point and field data, although usually separately from one another. Lastly, a survey by Jain~\cite{Jain:2010} discusses well known clustering techniques as well as the challenges involved in designing clustering algorithms.

%field features
We can also explore the specific techniques that already exist for segmentation and feature extraction in field-based datasets. One popular set of techniques involve region growing. Huang et al.~\cite{Huang:2007} used an interactive region growing method coupled with various morphological operations in order to extract multi-scale features from volume data. Additionally, Pra{\ss}ni et al.~\cite{Prassni:2010} utilized a uncertainty based region growing technique to minimize extraction errors that may falsely mask structures of interest or include unwanted background fluctuations. Furthermore, Ram{\'i}rez et al.~\cite{Ramirez:2013} took the GrabCut algorithm~\cite{Boykov:2001}, an image segmentation technique that treats pixels as a flow network, and extended it to 3D volumes.

An alternate method by Bremer et al~\cite{Bremer:2011} used the underlying topology to construct hierarchical merger tree of contours formed at varying data values. This is then used as a highly compact representation of features within the data. Another hierarchical approach was done by Ip et al.~\cite{Ip:2012} who used an intensity gradient histogram to continually subdivide a volume into a hierarchal representation of spatial structure that users can traverse and explore. When it comes to time-varying data and feature tracking, Ji et al.~\cite{Ji:2003} used an approach that constructs isosurfaces in higher dimensions (4D and 5D) to better compute and represent the temporal evolution of 3D volume features. Similarly, Weber et al.~\cite{Weber:2011} constructed a 4D Reeb graph to track volume surfaces features over time. Lastly, Balabanian et al.~\cite{Balabanian:2008} used temporal compositing and time-varying transfer functions to create a 4D multi-volume raycaster which can depict the evolution of volume features over time.

%point features
There has also been extensive work done in feature extraction and segmentation for point data, specifically for trajectories. Li et al.~\cite{Li:2013} developed a streamline similarity metric using a spatially sensitive bag-of-features approach and apply their technique to streamline querying and clustering. They also extend their work to develop a user driven streamline segmentation technique which can operate on multi-scale features~\cite{Li:2015}. Furthermore, Alewijnse et al.~\cite{Alewijnse:2014} used start-stop matrices and criteria-based segmentation to subdivide trajectories into various behaviors and applied their results to movement data. In addition, Ferstl et al.~\cite{Ferstl:2016} recently developed a method of visualizing the uncertainty of ensemble flow fields by clustering streamlines in a principal component analysis space. Confidence ellipses are constructed and projected back into a physical space to represent the variability of the ensemble.

In terms of more visual based analyses, Marchesin et al.~\cite{Marchesin:2010} devised a method of placing and selecting streamlines based on both the properties of a flow and the user viewing angle. These two selection criteria helped to improve the readability of normally cluttered streamline visualizations. In addition, Andrienko et al.~\cite{Andrienko:2009} developed an interactive interface to enable user-driven clustering of large trajectory datasets based on visual trends and Schrek et al.~\cite{Schrek:2009} used self-organizing maps to enable users to supervise and control aspects of trajectory clustering. 

When it comes to non-trajectory based point data, Linsen et al.~\cite{Linsen:2008} used 3D star coordinates to cluster and extract surfaces from multivariate particle data. These surfaces are able to segment the data with respect to an underlying multivariate function. Additionally, Pauly et al.~\cite{Pauly:2003} developed a multi-scale classification operator that uses principal component analysis in local neighborhoods to detect and extract features from unstructured point clouds. This was then applied to point sampled surfaces to generate line-based renderings of the extracted feature curves.

%field + point together
While limited, there has also been some work which tried to utilize both the point and field-based representations simultaneously. Sauer et al.~\cite{Sauer:2014} developed a method of tracking volume features at discrete jumps in time. While the work represents previously extracted volume features as groups of particles for the purpose of tracking, it does not incorporate any feature detection or segmentation. Additionally, J{\"o}nsson et al.~\cite{Jonsson:2009} used both representations by utilizing a point-based advection method to propagate field samples to interpolate regions with missing data in ocean satellite data. Furthermore, Agranovsky et al.~\cite{Agranovsky:2014} used in situ generated trajectories to more accurately store and represent vector field information for post hoc analysis. Other notable works include that by Salzbrunn et al.~\cite{Salzbrunn:2008} who used pathline predicates to define structures in unsteady flows and Chandler et al.~\cite{Chandler:2013} which used illustrative volume rendering to visualize time-varying properties of particle data.

%SLIC
Lastly, there is very notable and relevant work done in image analysis and computer vision. Specifically, Achanta et al.~\cite{Achanta:2010,Achanta:2012} developed an image clustering technique called Simple Linear Iterative Clustering (SLIC). This technique is based on a $k$-means clustering approach, but with two main distinctions. First, the search area is limited to regions near cluster centers making the algorithm extremely efficient and parallelizeable. Second, it uses a weighted distance metric which allows for easy control over the size and compactness of clusters.

Since then, Xie et al.~\cite{Xie:2015} have gone on to extend the SLIC algorithm to operate on 3D volumes and use ``supervoxels'' in conjunction with uncertainty-based refinement to extract volume features from large-scale datasets. Such a technique has a great deal of potential, especially if it can be applied in 4D and extended to operate on both field and point-based data types. As a result, we use the ideas originally established in the SLIC algorithm as a starting point in developing our multifaceted 4D feature segmentation and extraction scheme.

% METHODS %
%%%%%%%%%%%%%%%%%%%%%%%%%%%%%%%%%%%%%%%%%%%%%%%%%%
%%%%%%%%%%%%%%%%%%%%%%%%%%%%%%%%%%%%%%%%%%%%%%%%%%

\begin{figure}[t]
 \centering
 \includegraphics[width=0.99\linewidth]{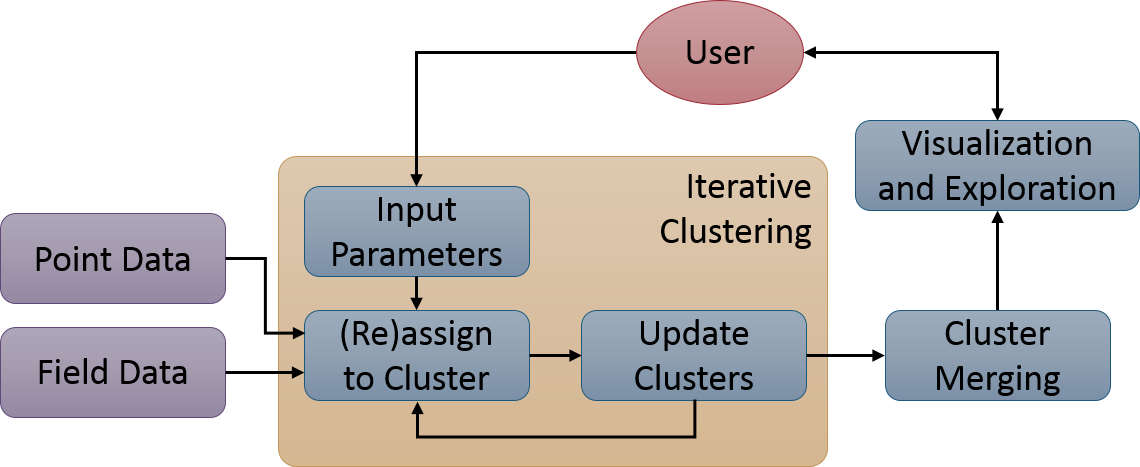}
 \caption{An overview of a typical workflow using this technique. Point and field-based data are collectively segmented using an iterative clustering algorithm. A user can visualize and explore the resulting features as well as adjust key clustering parameters.}
 \label{fig_workflow}
\end{figure}

\begin{figure*}[t]
 \centering
 \includegraphics[width=0.7\linewidth]{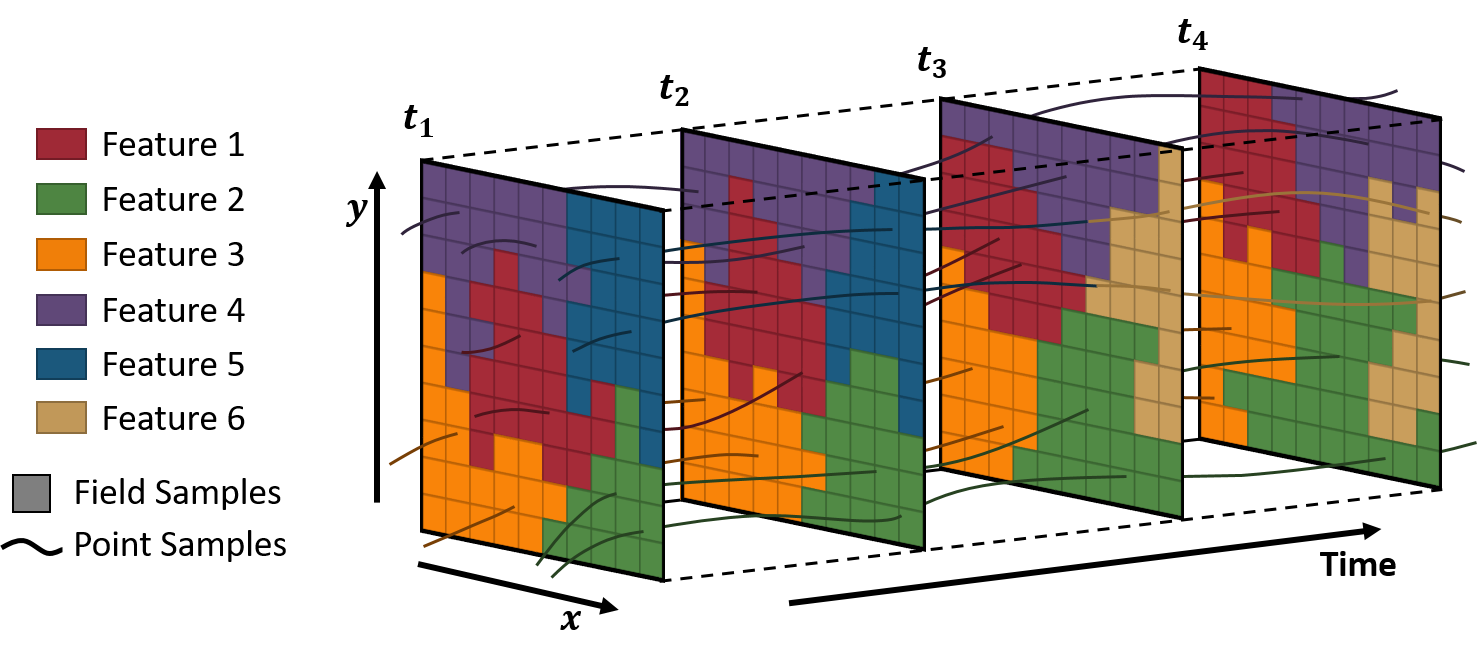}
 \caption{A 3D (2 space + 1 time) depiction of the multifaceted segmentation technique. Both field cells and point data are clustered based on their location and associated variables. The result is a set of meaningful spatio-temporal features (each shown in a different color) that can be isolated for further exploration. In this example, features 1-4 exist at all four timesteps, whereas features 5 and 6 exist at only the first and last two timesteps respectively. While a 3D version is depicted here for ease of illustration, the technique itself is able to operate in 4D (3 space + 1 time).}
 \label{fig_segmentation}
\end{figure*}

\section{Methods}

As previously described, we start by leveraging the ideas from the SLIC algorithm in order to be able to segment both point and field-based data simultaneously and in an efficient manner. Furthermore, we extend it to operate directly in 4D so that resulting features are representative of both spatial and temporal patterns found in the data. Figure~\ref{fig_workflow} shows an overview of workflow steps in this scheme. The point and field-based data are simultaneously segmented using our expanded version of the iterative clustering algorithm. The resulting segmentation produces a set of 4D multifaceted features that users can visualize and explore in both space and time. Adjusting key parameters in the iterative clustering scheme gives users control over the types of features produced and provides an additional means of interacting with the data.

\subsection{Data Representation}

In this description, the point-based data is collection of points whose position and other associated value(s) vary over time. A point $P$ can therefore be represented as:
$$P(t) = \{x_p(t), y_p(t), z_p(t), v_p(t)\}$$
\noindent In this case, $v_p$ is a variable associated with the point and can represent certain associated properties, such as momentum, mass, or age. Furthermore, the point data can often be formed into a set of trajectories allowing one to also consider their geometric properties. Some of these properties, such as curvature, vary along a trajectory and can be mapped to each point individually, while others, such as the total number of turns, represent the trajectory as a whole.

On the other hand, the field-based data is a collection of grid locations whose position do not change. As a result, a grid location $F$ can be represented as:
$$F(t) = \{x_f, y_f, z_f, v_f(t)\}$$
\noindent Since variables associated with the grid location do change, $v_f$ is still time-varying. Another difference is the structure that can be found within field-based representations as each grid location has a distinct set of neighbors. This provides a convenient means of both traversing and subdividing the domain.

Since the iterative clustering scheme only searches local regions around each cluster, we need a fast way to connect point and field samples with specific cluster centers based on spatial proximity. This will allow the system to quickly identify which clusters a particular sample needs to be compared to. We use a preprocessing step to generate a list of indexes that link each field location with a group of spatially associated point samples (i.e. which point samples lie within the volumetric space that is covered by a particular field sample). This helps to provide spatial organization to the point-based data. Furthermore, determining which clusters a certain field sample needs to be compared to, will automatically determine the same information for its associated point samples.

\subsection{Iterative Clustering} \label{sec:iterative_clustering}

In the original SLIC image segmentation algorithm, pixels are first assigned to an initial clustering which evenly divides the image spatially. For each cluster, a ``cluster center'' is generated by computing the average spatial location and color of all pixels in the cluster. In each iteration of the algorithm, pixels are compared to nearby clusters via a weighted distance metric that incorporates the spatial distance and color distance to the compared cluster center. Once all pixels have been assigned to a new (or the same) cluster, cluster centers are recomputed. This repeats until the algorithm converges onto a final clustering. More detailed information can be found in~\cite{Achanta:2010,Achanta:2012}. Our new technique however, must operate on 4D point and field-based data requiring significant changes and extensions to the original SLIC algorithm. 

Since point and field data often record a variety of different values, we first allow users to choose desired variables of interest. Spatio-temporal patterns in these variables will be used by the algorithm to segment the data into a set of 4D features which contain both a point and field-based counterpart. Furthermore, users can choose either a raw variable from each data type or form derived variables from groups of raw variables. These selections will determine what values to use for $v_p(t)$ and $v_f(t)$ for each data type. Figure~\ref{fig_segmentation} depicts an example of what the segmentation would look like for a time-varying dataset containing point and field samples.

\begin{figure*}[t]
 \centering
 \includegraphics[width=0.9\linewidth]{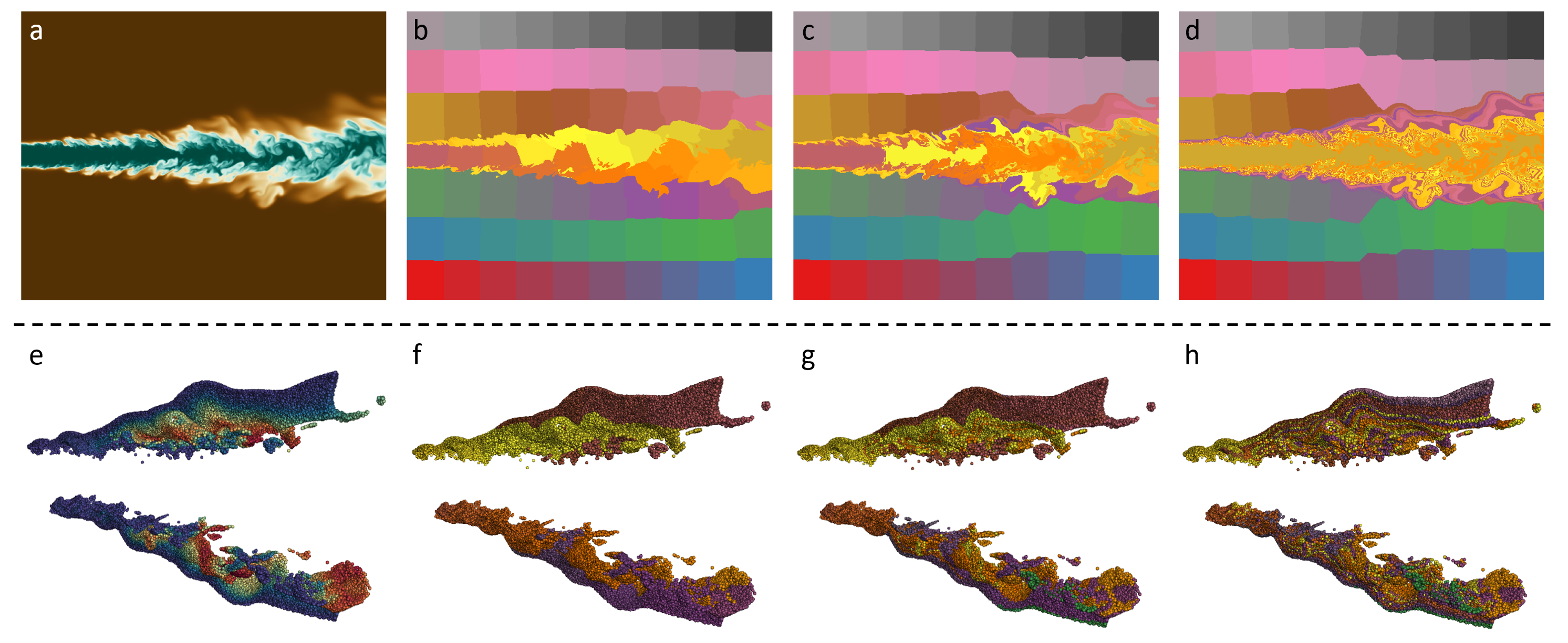}
 \caption{An image demonstrating the effect of adjusting distance metric weights in a combustion dataset. a) A 2D slice of the volume data. A jet of fuel (green) mixes with a background oxidizer (brown). b-d) Decreasing $w_d$ forms less spatially compact clusters that adhere more closely to fuel/oxidizer fluctuations. Each color represents a distinct cluster. e) A subset of the particle data. Warm (red) colors indicate a high concentration of hydroxide. f-h) Decreasing $w_d$ forms long/thin clusters that adhere closely to a constant hydroxide concentration.}
 \label{fig_weights}
\end{figure*}

\subsubsection{Cluster Initialization}

After deciding the point and field-based variables, users then choose the number of desired clusters, whose centers will be seeded initially in a 4D rectilinear grid fashion. The user determines how many cluster centers to seed in each of the four dimensions. This will help to account for differences in domain extent and resolution in each direction. If the full extent of the domain in dimension $i$ is $E_i$ and the number of cluster centers in that direction is $k_i$, then the interval distance between clusters is $C_i = E_i/k_i$. In our 4D case, $C_1$, $C_2$, and $C_3$ are spatial distances, and $C_4$ is a temporal distance.

Once the initial cluster centers have been seeded, all point and field samples over all timesteps are assigned to the nearest cluster center to form an initial set of clusters. However, care must be taken when computing the space-time separation between each object. This is because while the spatial unit for the first three dimensions is consistent (e.g., meters), the temporal unit for the fourth dimension is different (e.g., seconds). We therefore apply a user adjustable conversion factor, $c_f$, to transform the temporal unit to match the spatial units. In other words, $c_f$ could let a user define how many meters correspond to one second. Once all the units match, the space-time separation can be computed using a 4D Euclidean distance:
$$S_{st} = \sqrt{(x_c - x_s)^2 + (y_c - y_s)^2 + (z_c - z_s)^2 + (c_f(t_c - t_s))^2}$$
where subscript $c$ indicates the position/time of the cluster center and subscript $s$ indicates the position/time of the point or field sample. Note that this conversion factor also provides control over the size of resulting features in time compared to space. This is because a large value will make timesteps seem farther apart and a small value will make timesteps seem closer together.

\subsubsection{Iteration Loop}

Once the initial clusters have been generated, the iterative portion of the algorithm begins. First, new cluster centers are computed using the point and field samples in each cluster. These cluster centers contain six values: a space-time location representing the average location of all samples ($x_c$, $y_c$, $z_c$, and $t_c$), a point variable representing the average of all point sample values ($p_c$), and a field variable representing the average of all field sample values ($f_c$). Overall, each cluster center is represented as:
$$C_c = \{x_c, y_c, z_c, t_c, p_c, f_c\}$$
Once the new cluster centers have been computed, point and field samples can be compared to the cluster centers via a distance metric, and minimizing this metric will determine which cluster to assign each of the samples.

Similar to the original SLIC algorithm, we only search within a local area around each cluster center, so that each sample is only compared to a few cluster centers. This speeds up the algorithm since the total number of clusters can become quite large in a 4D space which may need a high cluster center resolution to segment large datasets rich in small features. Since the expected size of the clusters is approximately $C_1 \times C_2 \times C_3 \times C_4$, we search in a region of size $2C_1 \times 2C_2 \times 2C_3 \times 2C_4$ around each of the cluster centers.

Since the cluster centers change (in location and value) with every iteration, so will the samples that are assigned to it. This procedure repeats until all the cluster centers stop changing (the percent change of all six cluster center values lie below a user-defined error threshold, $\epsilon_c$). User control over the error threshold, $\epsilon_c$, provides balance between the performance and accuracy of the iterative clustering. In practice, we find the algorithm nearly always converges in under ten iterations.

\subsubsection{Distance Metrics}

Another major difference from SLIC is that we need to incorporate two entirely different types of samples (point and field values) into our cluster. Since each type of sample has its own variable being considered, comparing it to a cluster center is less straightforward. In SLIC only one type of sample was considered, as each sample (a pixel) contained a color and was compared to the color of cluster centers. In our case, we need a separate distance metric for point and field samples and a means of comparing each to cluster centers which contain information about both data types.

For comparing the point samples to cluster centers, we use a weighted linear combination of the difference in value and the difference in 4D position. Since each point sample only contains its own variable, $p_s$, we can only compare it to the point value average of the cluster center, $p_c$:
$$ D_p = w_p|p_s - p_c| + w_d S_{st}$$
We can use a similar distance metric for comparing the field samples to cluster centers. As with the point samples, since each field sample only contains its own variable, $f_s$, we can only compare it to the field value average of the cluster center, $f_c$:
$$ D_f = w_f|f_s - f_c| + w_d S_{st}$$
Using these two different distance metrics for each sample type will still be able to produce locally coherent features since the space-time separation term, $w_d S_{st}$ is common to both.

The three different weights in these distance metrics can be used to influence the size and compactness of the point and field-based portions of the resulting 4D features. Since $w_d$ is part of both metrics, it will adjust the feature as a whole. For example, setting the weight large will put more emphasis on space-time distance between samples and cluster centers resulting in compact features that allow for more variation in the values of its contained samples. On the other hand, setting the weight small will put less emphasis on the space-time distance resulting in larger and more spread out features that have very little variation in the values of its contained samples. The other two weights, $w_p$ and $w_f$, can be used to control the point and field-based portions of the features independently. For example, a user may wish to extract features that have a more compact (but internally varying in value) field-based portion and a less compact (but internally constant in value) point-based portion. Figure~\ref{fig_weights} shows an example of how adjusting distance metric weights can affect the clustering result.

\begin{figure}[t]
 \centering
 \includegraphics[width=0.9\linewidth]{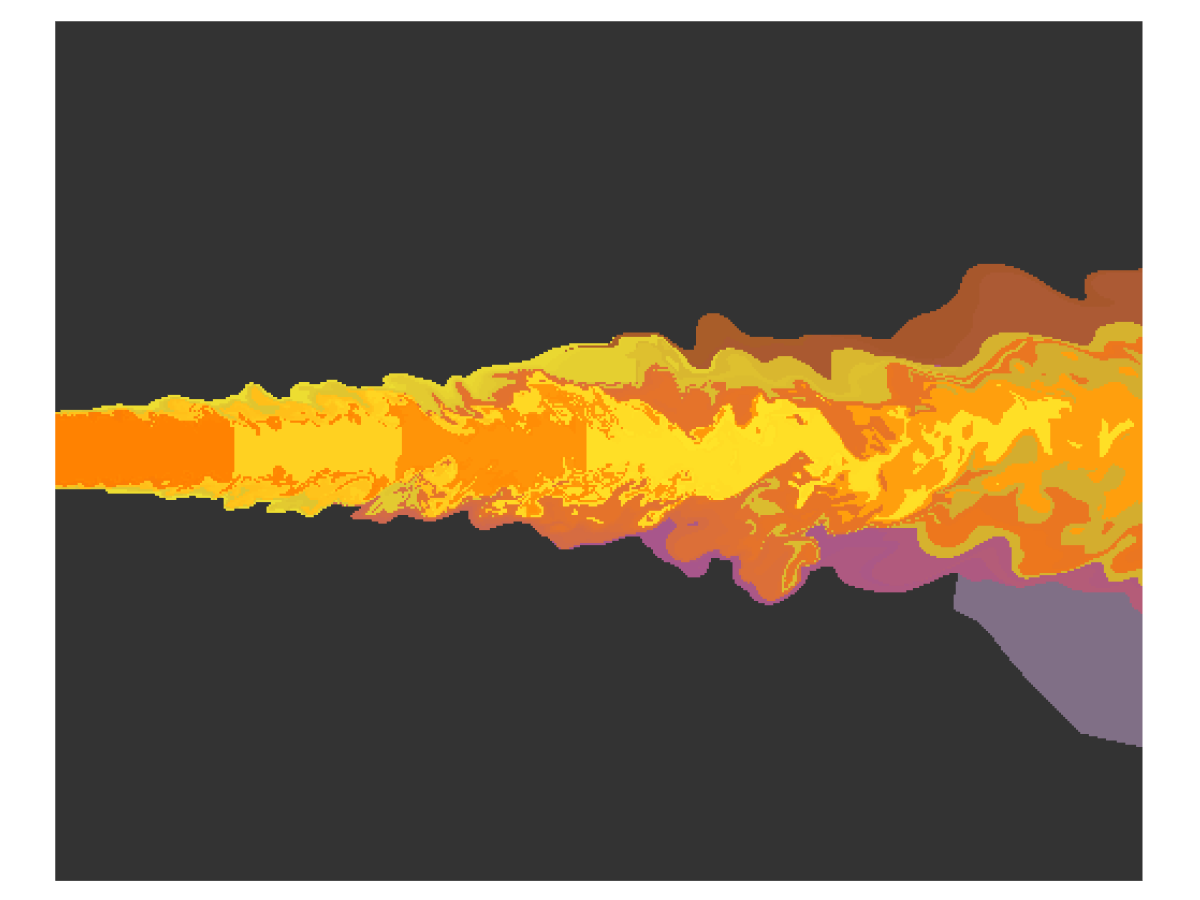}
 \caption{An image similar to those in Figure~\ref{fig_weights}b-d, but with background clusters merged into a single cluster (drawn in black). This cluster can then be ignored as it contains no interesting data fluctuations in either the point or field data types. The single gray-purple feature towards the bottom right is not merged into the background because, although it has no interesting field patterns, there are some fluctuations in the particle portion of that feature (not shown).}
 \label{fig_merge}
\end{figure}

\subsection{Cluster Merging}

Since this technique uses a k-means clustering approach, it is possible to over-segment or under-segment the data depending on how many initial clusters are chosen. We recommend over-estimating the number of clusters especially because the complexity of the algorithm is dependent primarily on the number of point/field samples instead of the number of clusters. Then, any over-segmentation caused can be repaired by merging clusters that have similar cluster centers. This is especially useful for portions of the domain, such as the background, where no interesting patterns are emerging (as in Figure~\ref{fig_weights}b-d). Merging this region into a single ``supercluster" will allow a user to easily discard unimportant data points.

Note that the clustering algorithm itself does not enforce any spatio-temporal connectivity and allows features (clusters) to consist of multiple connected components. This can be useful in certain cases, such as extracting exact boundaries between varying variable values (by setting $w_d$ to be small) in turbulent portions of a domain. Similarly, we do not enforce any spatio-temporal connectivity when merging similar features and ignore the four space-time location terms in the cluster centers. We merge any features whose cluster centers are very close in value (the percent difference between the values in the two centers lie below a user-defined error threshold, $\epsilon_m$). Note that we only merge if both the field and point averages match one another, since two features that have similar centers in one data type but not the other should still be treated distinctly from one another as they can represent entirely different phenomena. User control over the error threshold, $\epsilon_m$, ultimately determines how much merging will occur. Figure~\ref{fig_merge} shows an example of merging clusters to remove the uninteresting background.

%After the cluster merging step is complete, features are ready for visualization and exploration. A summary of all of the steps can be found in... (algorithm 1).
%Algorithm steps:
%User decides variables of interest (one field and one trajectory)
%User decides number of clusters (in xyz and t directions)
%User decides all weights (variable weights, proximity weight, temporal conversion)
%Seed and construct initial clusters (regular grid)
%Iteration begins:
%-Within search areas, assign each voxel to most similar cluster
%-Update cluster centers (4d position + average of all elements)
%-Keep going until cluster centers stop changing
%Merging

\begin{figure*}[t]
 \centering
 \includegraphics[width=0.9\linewidth]{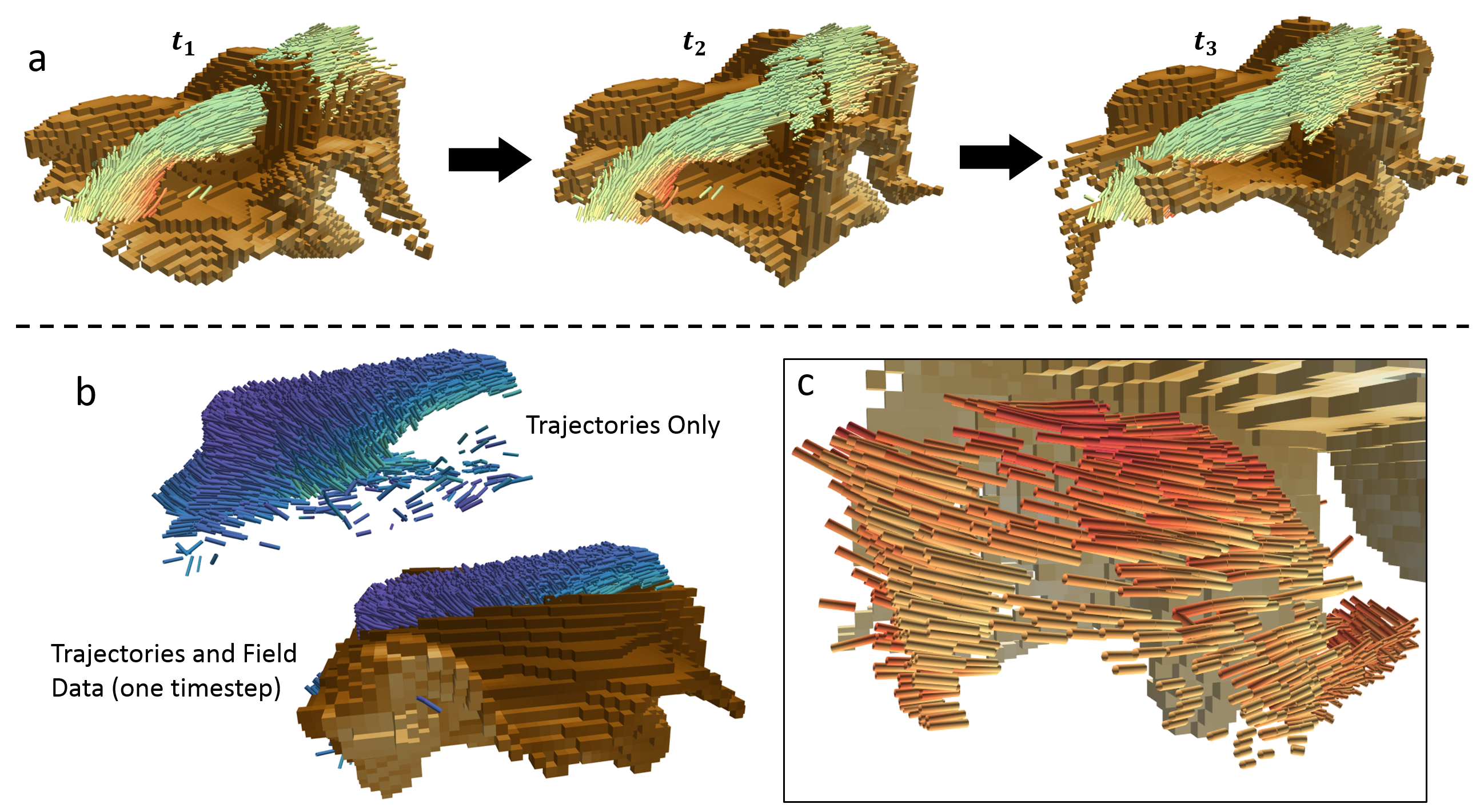}
 \caption{An image showing various multifaceted features extracted from the combustion dataset. The field data represents the mixture fraction variable while the particles represent hydroxide concentration and use the same color mapping as in Figure~\ref{fig_weights}a/e. a) Particle trajectories shown with three select timesteps of the volume portion of a feature transitioning into a burning state. b) A feature where no burning is occurring showing the trajectory portion only (top left) and both the trajectories and a volume timestep (bottom right). c) A portion of a feature where burning is occurring. Trajectories are wrapping around the volume portion of the feature. For more information see Section~\ref{sec:combustion}.}
 \label{fig_combustion}
\end{figure*}

\subsection{Implementation and GPU acceleration} %make gpu a subsection?

One of the major advantages of this technique is that each point or field sample can be treated independently from one another during each clustering iteration. More specifically, computing the distances to cluster centers and assigning the sample to a cluster can be done simultaneously for all samples. This type of processing lends itself extremely well to GPU acceleration. In our implementation, cluster center information resides in GPU global memory. Each GPU thread computes the distance metrics for a particular sample and assigns it to the appropriate cluster. Furthermore, the new cluster centers can be updated on-the-fly by using atomic operations to keep a running average of all samples placed in the cluster so far. Once all samples have been processed, the new cluster centers are already available for the next iteration, never having to leave GPU memory.

The main challenge with such an approach is the limited amount of available GPU memory as a typical dataset is much too large to fit in its entirety. To circumvent this, one could either use multiple GPUs in a distributed setting or using a single GPU that processes samples in several chunks. Note that if multiple GPUs are being used, then cluster center information must be communicated and merged after each iteration. In our current implementation, we use a single GPU in order to demonstrate that this technique can handle large scientific datasets using only a standard desktop PC.

Another advantage is that since samples can be treated as independent 4D objects, the point and field datasets do not need to have matching temporal resolutions. This is often the case when dealing with multifaceted data, such as the test datasets used in this paper, which have different temporal resolutions in their point and field counterparts. Furthermore, this technique can also be used for non-multifaceted datasets that have only a point-based or field-based component. In such a case, there would be one type of sample which uses only one of the distance metrics.

After the segmentation is complete, potentially thousands of features can be produced. Users need a way to guide their selection process to determine which features to explore in more detail. We find that the final cluster center information gives a good overview of the types of features that have been extracted (in terms of point and field variables) and where in the domain they are located. We populate a list of cluster centers and allow users to choose features of interest to visualize and explore in more detail. The size of this list can be reduced through range queries of cluster center properties. Furthermore, other information can be computed and displayed as well, such as the spatio-temporal extent and the standard deviations of the contained point and field samples.

It is very difficult to intuitively visually represent the resulting 4D multifaceted features in their entirety. This is because visual clutter and occlusion limit what can be shown from the 4D feature space. The point-based component is more straightforward since it can be rendered as a set of trajectories, an already intuitive 4D object. Furthermore, clutter from too many trajectories can be reduced by only drawing a temporal subset of the full trajectory length in the feature. We use shaded path tubes to help give users a better sense of the 3D structure. Note that it is possible that some points from a trajectory become assigned to a different feature. When this occurs, we choose to split the trajectory into separate components and only connect points where there are at least two consecutive timesteps as part of the feature.

Alternatively, the field-based component tends to consist of several large overlapping volumetric regions and representing them all at once would result in too much clutter making it difficult to understand the spatio-temporal evolution. As a result, we display only one timestep at a time and allow users to seek through all available timesteps to visualize the evolution of the field data. In our implementation, we render the boundary of each volumetric surface using cubes to make individual voxels clear from one another, however a smooth boundary surface could be used as well. These surfaces are displayed in conjunction with the path tube trajectories to see how the two data types interact with one another in space and time. Note that there may be alternate ways of visually representing these types of features and will be explored more in future work (see Section~\ref{sec:discussion}).

%figure for visual representation (splitting trajectories, volume surfaces)???

% RESULTS %
%%%%%%%%%%%%%%%%%%%%%%%%%%%%%%%%%%%%%%%%%%%%%%%%%%
%%%%%%%%%%%%%%%%%%%%%%%%%%%%%%%%%%%%%%%%%%%%%%%%%%

\section{Results}

We test this technique using two real world multifaceted datasets that have both a point/trajectory and field-based component. The first dataset is a scientific simulation dataset that comes from a large-scale combustion simulation. This simulation represents and saves time-varying data as both tracer particles and on a 3D volume. The second dataset is a city-based dataset representing geospatial movement data from taxis (a point-based representation) as well as traffic patterns at various locations (a field-based representation). This second dataset shows that our technique has a broader applicability outside that of standard scientific simulation data since there are many other types of multifaceted datasets that contain both point and field-based components. We use case studies to demonstrate the types of features that can be extracted and provide performance results to show that these types of datasets can be handled using a standard desktop PC.

\begin{figure*}[t]
 \centering
 \includegraphics[width=0.9\linewidth]{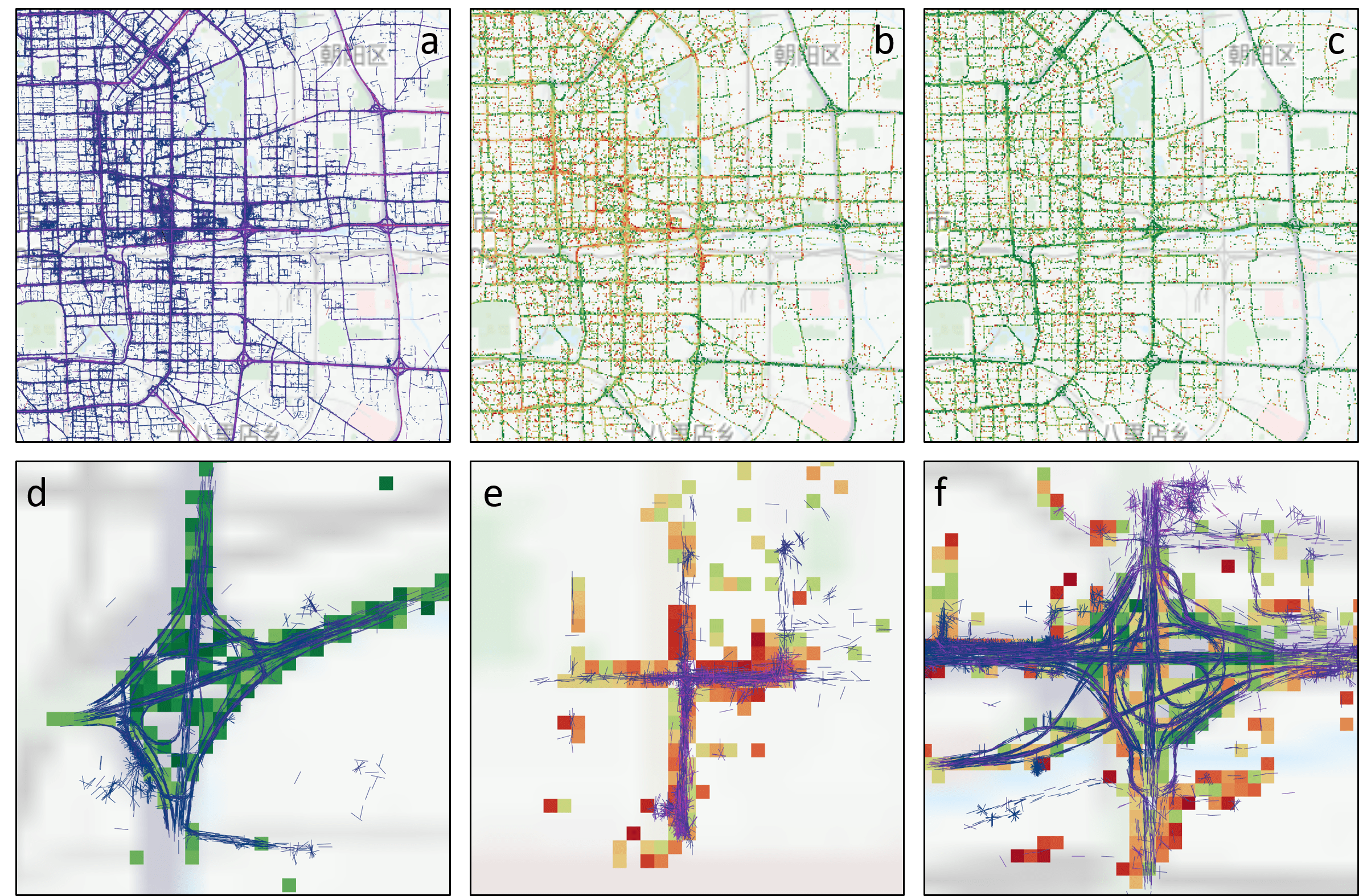}
 \caption{Images from the city dataset. a) One day's worth of trajectory data in the eastern part of the city colored according to speed. Blue points represent slower vehicles while purple/pink points represent faster vehicles. b/c) Traffic levels at 6pm (rush hour) and 9pm (after rush hour) respectively. Low traffic areas are shown in green while high traffic areas are shown in orange/red. d) A feature extracted from the data representing a low traffic highway intersection. The blue trajectories indicate taxis taking an indirect route to their destination. e) A feature representing a high traffic local city intersection. The purple/pink trajectories indicate taxis taking a direct route to their destination. f) A feature representing a different highway intersection which contains both high and low traffic regions as well trajectories taking both a direct and indirect route. For more information see Section~\ref{sec:city}.}
 \label{fig_city}
\end{figure*}

\subsection{Combustion Dataset} \label{sec:combustion}

The combustion dataset comes from S3D~\cite{Yoo:2011} a large-scale simulation developed by researchers at Sandia National Laboratories. Results from this simulation are used to study various combustion processes and are crucial towards designing more efficient engines. The particular simulation run we use represents a turbulent lifted ethylene jet flame and contains time-varying data in the form of tracer particles and a 3D volume. Researchers are interested in events leading up to and after combustion, when fuel and oxidizer mix in just the right amounts in order for burning to occur. 

The field-based portion of this dataset represents a 3D volume containing over 20 million field samples per timestep. The variable we choose to look at is the "mixture fraction'' variable, which measures the ratio of fuel to oxidizer in various parts of the domain. Figure~\ref{fig_weights}a shows 2D slice of the volume with fuel drawn in green and oxidizer drawn in brown. Light beige areas indicate where the two substances are mixing. The point-based portion consists of over 1 million massless tracer particles per timestep, which measure the mass fractions of various chemical substances. The variable we choose to look at is the mass fraction of hydroxide (OH\textsuperscript{-}) a direct byproduct of the combustion process that can indicate where burning has occurred. Figure~\ref{fig_weights}e shows a subset of the particle data with warmer colors (orange/red) indicating a higher hydroxide concentration and cooler colors (blue/purple) indicating a lower hydroxide concentration. Note that although the particle data contains fewer samples per timestep it is saved at a much higher temporal frequency.

We processed this dataset using our segmentation technique and produced approximately 1600 features. Distance metric weights were chosen to prioritize more spatial compactness on the particle data (instead of a constant variable value). This is because values can easily change along a trajectory and we wanted to ensure that particles belonging to the same trajectories would be more likely to end up in the same cluster. We reversed this constraint in the field portion allowing more variation in size and less variation in its variable values. This produced more amorphous and spread out volumetric regions with less variation in its variable values.

Figure~\ref{fig_combustion} shows various examples of the different types of features produced. Figure~\ref{fig_combustion}a shows the trajectory portion of the feature as well as three select timesteps of the volume portion. This likely represents a transitional region where burning is beginning to occur. The top portions of the feature contain green trajectories, indicating a small amount of hydroxide, and dark brown volume cells, indicating a region with mostly oxidizer. While some burning may be beginning to occur here, the ratio of fuel to oxidizer is not optimal. However, towards the bottom of the feature, we can see more reddish trajectories, indicating a higher amount of hydroxide, as well as more beige volume cells, indicating a more even fuel and oxidizer mixture. From the temporal evolution, we can see that this transitional region is slowly moving in unison towards the right in each image.

Figure~\ref{fig_combustion}b shows an alternate feature with the trajectories only on the top-left and the trajectories with a timestep of volume data on the bottom-right. The blueish purple color on the trajectories indicate that there is little to no hydroxide present in this feature. Furthermore, the majority of the volume portion is a dark brown color, indicating regions with mostly oxidizer. The beige portion near the bottom left does indicate a more even mixture between the fuel and oxidizer. However, the lack of hydroxide suggests that only mixing is occurring at the moment and that the region has not combusted yet. Lastly, Figure~\ref{fig_combustion}c shows a beige volume region with orange/red trajectories that is undergoing burning. An interesting pattern is occurring in that the trajectory component of this extracted feature seems to be wrapping around the volume component in a vortex-like fashion.

\subsection{City Dataset} \label{sec:city}

The city dataset is provided by the Computational Sensing Lab at Tsinghua University in Beijing~\cite{Yu:2010,Zhang:2011}. In this case study, we focus on the city of Beijing and couple the movement trajectory data of taxis (point-based data) with traffic patterns measured at fixed spatial locations (field-based data). City data like this is extremely useful for studying traffic and can aid in planning future transportation projects. Furthermore, analyzing the behavior of vehicles in various traffic conditions can help researchers better understand potential causes and solutions for road congestion.

The point/trajectory-based data consists of geospatial movement data from around 28,000 taxis taken over the course of an entire month with data points sampled at 1 minute intervals. This helps to provide both a high resolution description of an individual taxi's path throughout the city as well as long term fluctuations in traffic patterns in the city as a whole. In addition to geospatial location over time, other parameters are recorded as well, such as the speed of the vehicle, the direction it's facing, and whether or not it is carrying a passenger. However, it is the overall shape of the trajectory that becomes very important as it can indicate the behavior and decisions made by an experienced driver. Figure~\ref{fig_city}a shows an image of one day's worth of trajectory data in the eastern part of the city. In this image, trajectories are colored according to speed, showing slower data points in blue (such as those on local streets) and faster data points in purple/pink (such as those on freeways).

The field-based data represents traffic fluctuations throughout the city and is recorded on around 1 million grid locations per timestep. This traffic information is originally derived from the speed and motion of city vehicles at certain times throughout the day. However, it is recorded on a field-based representation because it is important to be able to analyze changes in traffic at fixed points of interest (major intersections, freeway mergers, bridges, etc.). Figure~\ref{fig_city}b and c show a difference in the traffic patterns at 6pm and 9pm respectively. Since 6pm is in the middle of rush hour, we can see many high traffic areas (shown in red) as workers head home for the day. However, the traffic levels reduce significantly (shown in green) just three hours later once rush hour has ended.

We can apply our multifaceted feature segmentation approach. However, instead of the 4D version of the algorithm as described previously, we use a 3D version with only two spatial coordinates plus time. In the field data, we used the level of traffic as described in the previous paragraph for the input variable. In the point data however, we derived a new variable based on the underlying shape of the trajectory. By taking the ratio between the displacement and overall path length of a taxi trip, one can estimate whether a driver chose to take a more roundabout path to their destination. In cases where there is a large discrepancy between the path length of the trip and the distance between the source and destination points, drivers may have taken an alternate route to avoid congestion.

Figure~\ref{fig_city}d-f shows examples of some of the multifaceted features extracted from this dataset. The field-based traffic levels are drawn as in the previous two images. However, the trajectory data is colored according to how direct a path the driver chose to take to their destination. Trajectories taking an indirect path are shown in blue while trajectories taking a more direct path are shown in purple/pink. Furthermore, trajectories are drawn as small disconnected line segments aligned with the direction that the vehicle is facing. Simply connecting the point data using lines can result in trajectories that do not follow the road network perfectly since a taxi can make a few turns within the 1 minute sampling resolution. 

All three of these features represent a time during the afternoon rush hour. Part (d) of the figure shows a feature representing a low traffic highway intersection. This feature only consists of blue trajectories which are taking an indirect route to their destination. One possible explanation is that the taxis are taking a roundabout path in order to pass through a less congested area. In such a case, taking an indirect path to avoid congestion takes the least amount of travel time. On the other hand, part (e) shows a high traffic local city intersection with primarily purple/pink trajectories which indicate that a driver is taking a more direct route to their destination. Sometimes taxis are forced to drive through a congested area because there are no other nearby clear routes. In other words, simply taking the most direct path through the congestion will take the least amount of travel time. Lastly, part (f) shows an alternate highway intersection which contains both high and low traffic regions as well as both types of trajectories.

%add more detail to last feature?

\begin{figure}[t]
 \centering
 \includegraphics[width=0.99\linewidth]{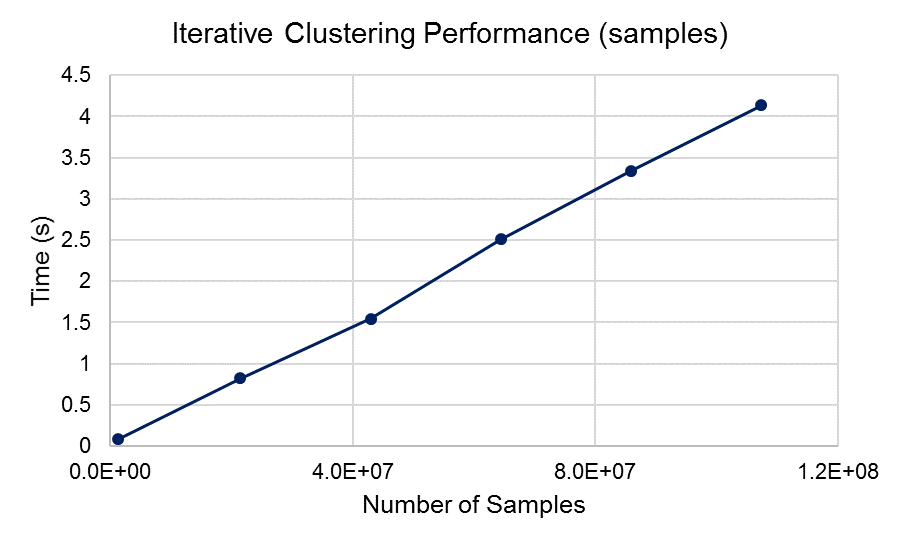}
 \caption{Performance results while varying the number of samples in a single clustering iteration. Results scale linearly with the number of samples due to limits in available GPU memory and aggregation required when computing new cluster centers.}
 \label{fig_perf_samples}
\end{figure}

\begin{figure}[t]
 \centering
 \includegraphics[width=0.99\linewidth]{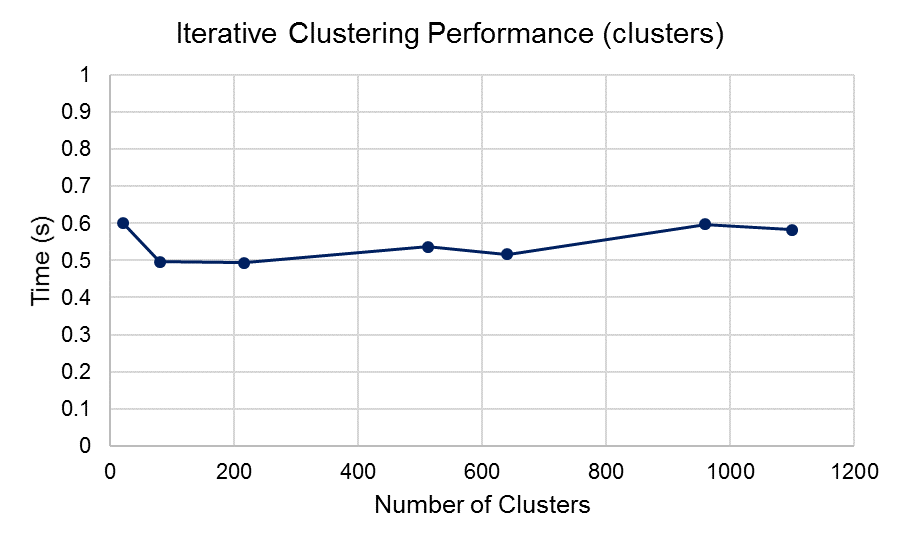}
 \caption{Performance results while varying the number of clusters in a single clustering iteration. Since samples are only compared to cluster centers in a local neighborhood, adjusting the number of clusters has little effect on the overall performance.}
 \label{fig_perf_clusters}
\end{figure}

\subsection{Performance Results}

Lastly, we provide some performance results to illustrate the computational time required to segment a 4D dataset. As previously described, this iterative clustering technique scales primarily with the number of samples that need to be clustered. However, the number of iterations required does vary between datasets and user chosen input parameters. We therefore provide performance results for one clustering iteration as this will be independent of the underlying data patterns found in different datasets. Note that in practice we find that the total the number of iterations required is usually quite small, generally resulting in convergence in under ten iterations. We use a standard desktop PC with an Intel i7-4790K CPU and a single Nvidia Geforce GTX 970 GPU for our implementation.

Figure~\ref{fig_perf_samples} shows that the performance of a clustering iteration tends to increase linearly with the number of samples. This is due to two main factors. First the data must be processed in chunks on the GPU due to available memory limitations. Several floating point values (xyz position, time, and variable values) need to be transferred to the GPU quickly taking up all available memory. As more samples are introduced, more chunks need to be sent to the GPU for processing. Secondly, computing the new cluster centers after each iteration requires aggregating the results of all samples within each cluster. Since samples cannot be treated independently in this final step, having more samples will also increase the computation time regardless of whether they all fit into GPU memory.

Figure~\ref{fig_perf_clusters} shows that varying the number of clusters has little effect on the performance of a clustering iteration. This was expected since the original SLIC algorithm also exhibits the same property. This is because samples are only compared to cluster centers within a local neighborhood that is dependent on the expected size of a cluster ($2C_1 \times 2C_2 \times 2C_3 \times 2C_4$ as described in Section~\ref{sec:iterative_clustering}). As the number of clusters increases and are seeded closer to one another, the local search area decreases proportionally. This ensures that samples do not need to be compared to cluster centers located far away either spatially or temporally as it is unlikely that those clusters will end up minimizing the distance metric.

These results show that a single desktop PC can not only segment large numbers of samples in a reasonable amount of time, but it can handle over a thousand clusters without significantly affecting clustering time. Furthermore, since this algorithm can produce such a large number of features simultaneously, it only needs to be run once as a preprocessing step. Upon completion, researchers can interactively explore the rich set of features produced for an extended period of time. However, when dealing with much larger data scales, a distributed computing environment with numerous GPUs is likely required. Samples can be split among computing nodes, significantly improving performance since each data subset is small enough to fit entirely within GPU memory. More tests are required to fully evaluate the scalability of this algorithm in a distributed system and will be investigated in the future.

% DISCUSSION %
%%%%%%%%%%%%%%%%%%%%%%%%%%%%%%%%%%%%%%%%%%%%%%%%%%
%%%%%%%%%%%%%%%%%%%%%%%%%%%%%%%%%%%%%%%%%%%%%%%%%%

\section{Discussion} \label{sec:discussion}

These results demonstrate the unique types of features that can be extracted using this multifaceted segmentation technique. By incorporating information from both a point and field-based data type, researchers can gain a more diverse understanding of underlying phenomena. However, there are many ways in which the ideas presented here can be expanded or applied to other venues. In this section, we discuss ways of understanding the meaning behind a particular extraction result, what other data types this technique is applicable towards, as well as the current limitations of this implementation and potential future directions to address them.

\subsection{Understanding the Extraction Result}

Subdividing a large complex dataset into a set of meaningful features is an extremely useful data reduction tool. Not only can it reduce the scale of the data, making it more manageable, but it allows users to focus their exploration in a more detailed fashion. While there are many methods available to explore the properties of a resulting feature, one can also make a separate set of interpretations about the data by understanding the reasoning behind why the algorithm ``chose" to cluster the data in this manner. In other words, investigating the complex spatio-temporal boundaries between segmented features can provide information about the dataset as a whole and how the different data representations interact with one another.

In general, this scheme tends to extract regions with similar variables in the point and field-based datasets in consistent spatio-temporal regions within the domain. This is achieved through the space-time separation term that is common to both distance metrics. One example is to look at the spatio-temporal extent of the point and field based components since they can depict correlations between the chosen variables in each data type. If the boundaries of the point-data and field data match, then there is likely a strong connection between the two variables in that part of the domain. On the other hand, if each component occupies its own distinct spatio-temporal extent, then the variables behave independently from one another. 

Furthermore, one can adjust the various input parameters, such as the number of clusters in each dimension, temporal conversion factor, or the distance metric weights, and explore how this changes the overall segmentation result. For example, adjusting the distance metric weights to prioritize the space-time separation or variable values by different amounts can highlight when and where data values are rapidly changing throughout the domain. This is because at some point, there will be a balance between the similarity between a sample and a cluster value and the overall space-time separation between the two. Adjusting these weights will affect where this balance lies. In addition, certain combinations of variables may be more or less sensitive than others when changing these parameters and can also significantly affect the extraction result.

\subsection{Applicable Data Types}

While this paper focuses primarily on spatio-temporal data that consists of both a field-based and point-based representation, there are other data types and formats that are applicable to this multifaceted segmentation technique. Firstly, both of the combustion and city-based data store the field values on a regular rectilinear grid. However, this technique can also apply to field data stored on unstructured grids as well with little modification to the algorithm itself; field samples can still be clustered using the same distance metric. The primary difference is that determining which nearby cluster centers need to be checked for each field sample becomes more computationally intensive. This is because now the field data and the cluster centers (which are free to move after each iteration) both lack any predictable spatial structure. One possible solution would be to use organizational techniques, such as a hierarchical space partitioning, to give structure to the data samples.

Another option is to forgo the physical space altogether. While a vast majority of scientific datasets try to represent phenomena from the real physical world (in a 3D or 2D spatial domain), it can be beneficial to also apply this type of segmentation to a more abstract phase space. In this case, a trajectory would represent the path that a point sample takes throughout the phase space over time. As for the field data, samples can be either projected into the phase space like the point samples, or a new grid-like representation could be defined. Once in this new space, both the point and field samples can be clustered as before given appropriate distance metrics. Enabling this type of multifaceted phase space segmentation would further generalize this feature extraction technique.

Lastly, one could apply this technique to other data representations (besides point-based and field-based data). Although many datasets use the two representations we focus on in this paper, there are other data types, such as network or tree-based data, that are used in various scientific applications. For example, many cosmology simulations construct merger trees that track the joining of dark matter halos over time due to gravitational attraction. Just like the point or field-based data types, this merger tree also exists in 4D space and has properties unique to that data type. Using this type of data as an additional input into our multifaceted segmentation technique could allow one to cluster segments and branches of the tree into various features. Such a resulting feature could potentially contain a point, field, and tree-based component all at the same time.

\subsection{Limitations and Future Work}

There are still some limitations associated with this technique, which in turn, open up several venues for future research. First, we find that the user defined input parameters (distance metric weights, temporal conversion factor, number of cluster centers to seed, etc.) need to be carefully chosen in order to produce a desirable segmentation result. Given a new dataset, it can take time to learn what values work well and how each will affect the types of features being produced. In the future, it may be possible to devise a scheme that can either suggest or automatically choose these input parameters based on the patterns and underlying structure of the input dataset.

Next, there are limits in the way these 4D multifaceted features are visually represented. Clutter and occlusion play a large role since the point/trajectory and field-based portions often overlap in space and time. Adding transparency can help in certain instances, but can lead to misleading interpretations when dealing with very complex shapes. It may be possible to use isosurfaces or volume rendering techniques, but those alone will not be able fully capture the 4D evolution of both data types. More research is required in developing new visual representations and abstractions to portray the complex 4D structure found within such a feature.

Lastly, the way in which features are extracted through time can be expanded. Currently, the system connects samples from multiple timesteps into a feature if their variable values are similar. However, in many cases a feature evolves not only in shape, but also in its internal values. An alternative could be to extract features whose internal values all increase or decrease by the same amount over time. In the current implementation, this can be achieved by adjusting the way features are merged when adjacent temporally. In fact, more general control over the way similar features are merged can provide users with more control over extraction results, such as the ability to merge features hierarchically and explore a detailed merger tree. By traversing through different levels in the tree, users could control the spatio-temporal detail/resolution of the segmentation result.

%conduct more extensive performance tests in a distributed multi-GPU setting

% CONCLUSION %
%%%%%%%%%%%%%%%%%%%%%%%%%%%%%%%%%%%%%%%%%%%%%%%%%%
%%%%%%%%%%%%%%%%%%%%%%%%%%%%%%%%%%%%%%%%%%%%%%%%%%

\section{Conclusion}

Overall, this work presents a new feature segmentation and extraction scheme for datasets that have both a point and field-based representation. Such a technique is able to construct a set of 4D multifaceted features based spatio-temporal trends found in both of these data types. This enables researchers to not only reduce datasets into meaningful subsets, but also explore the spatial and temporal interplay between the point and field-based representations allowing them to study underlying phenomena from new perspectives. Case studies using combustion simulation data as well as geospatial city data illustrate the types of features that can be extracted and demonstrate the broad applicability of this approach. Furthermore, we employ GPU acceleration in order to exploit the fact that data samples can be treated independently from one another throughout much of the algorithm. Performance results show that a standard desktop PC can handle typical scientific datasets and produce potentially thousands of features simultaneously. As the use of multifaceted data continues to grow in popularity, techniques like these will become even more essential in being able to explore all aspects of a particular system of study.

\acknowledgments{This research is sponsored in part by the U.S. Department of Energy through grant DE-SC0012610 and DE-SC0007443.}

\bibliographystyle{abbrv-doi}

\bibliography{template}
\end{document}